\documentclass[sigconf, nonacm]{acmart}
\setcopyright{none}
\makeatletter
\let\@authorsaddresses\@empty
\def\@copyrightpermission{\relax}

\def\ACM@cc@type{}
\def\ACM@cc@image{}
\def\ACM@badge@width{0pt}
\def\ACM@badge@height{0pt}

\makeatother

\usepackage{microtype}
\usepackage{graphicx}
\usepackage{subcaption}
\usepackage{booktabs}
\usepackage{comment}

\usepackage{amsmath}
\usepackage{mathtools}
\usepackage{enumitem}

\usepackage[capitalize,noabbrev]{cleveref}

\begin{document}

\title{ARC-TGI: Human-Validated Task Generators with Reasoning Chain Templates for ARC-AGI}

\author{Jens Lehmann}
\authornote{Work done outside of Amazon.}
\affiliation{%
  \institution{Dresden University of Technology}
  \institution{Amazon}
  \city{Dresden}
  \country{Germany}
}
\email{jens.lehmann@tu-dresden.de}

\author{Syeda Khushbakht}
\authornote{Both authors contributed equally to this research.}
\affiliation{%
  \institution{Dresden University of Technology}
  \city{Dresden}
  \country{Germany}}
\email{syeda_khushbakht.batool@tu-dresden.de}

\author{Nikoo Salehfard}
\authornotemark[2]
\affiliation{%
  \institution{TIB - Leibniz Information Centre}
  \city{Hanover}
  \country{Germany}}
\email{nikoo.salehfard@stud.uni-hannover.de}

\author{Nur A Zarin Nishat}
\affiliation{%
  \institution{TIB - Leibniz Information Centre}
  \city{Hanover}
  \country{Germany}}
\email{nur.nishat@tib.eu}

\author{Dhananjay Bhandiwad}
\affiliation{%
  \institution{Dresden University of Technology}
  \city{Dresden}
  \country{Germany}}
\email{dhananjay.bhandiwad@tu-dresden.de}

\author{Andrei Aioanei}
\affiliation{%
  \institution{TIB - Leibniz Information Centre}
  \city{Hanover}
  \country{Germany}}
\email{andrei.aioanei@tib.eu}

\author{Sahar Vahdati}
\affiliation{%
  \institution{TIB - Leibniz Information Centre}
    \institution{Leibniz University of Hanover}
  \city{Hanover}
  \country{Germany}}
\email{sahar.vahdati@tib.eu}

\renewcommand{\shortauthors}{Lehmann et al.}

\begin{abstract}
The Abstraction and Reasoning Corpus (ARC-AGI) probes few-shot abstraction and rule induction on small visual grids, but progress is difficult to measure on static collections of hand-authored puzzles due to overfitting, dataset leakage, and memorisation.
We introduce ARC-TGI (ARC Task Generators Inventory), an open-source framework for \emph{task-family generators}: compact Python programs that sample diverse ARC-AGI tasks while preserving a latent rule. ARC-TGI is built around a solver-facing representation: each generated task is paired with natural-language \emph{input} and \emph{transformation} reasoning chains and partially evaluated Python code implementing sampling, transformation, and episode construction.
Crucially, ARC-TGI supports \emph{task-level constraints} so that training examples collectively expose the variations needed to infer the underlying rule, a requirement for human-solvable ARC tasks that independent per-example sampling often fails to guarantee. 
All generators undergo human refinement and local verification to keep both grids and reasoning traces natural and consistent under variation.
We release 461 generators covering 180 ARC-Mini tasks, 215 ARC-AGI-1 tasks (200 train, 15 test), and 66 ARC-AGI-2 tasks (55 train, 11 test), enabling scalable dataset sampling and controlled benchmarking.

\end{abstract}


\maketitle

\section{Introduction}
The ability to \emph{acquire new skills from limited experience} and apply them to novel tasks remains a core challenge in AI.
The Abstraction and Reasoning Corpus (ARC) \cite{chollet2019}--and its ARC Prize variants, ARC-AGI\footnote{\url{https://arcprize.org/arc-agi}} (e.g., ARC-AGI-1/2) --is designed to probe this capability with small colored grids.
Each \emph{task} consists of several training input--output grid pairs and one or more test inputs for which the solver must produce the corresponding outputs.
Despite the visual simplicity of the medium, many tasks require abstraction over objects, relations, and compositional rules, and humans can often solve them from just a few examples.
A persistent obstacle is evaluation.
ARC-style benchmarks are small and largely static, making it difficult to 
(i) measure progress reliably under overfitting and dataset leakage, 
(ii) distinguish generalization from memorization of particular puzzles, and 
(iii) run controlled studies that vary one factor (e.g., grid size, colors, distractors) while keeping the underlying rule fixed.
A natural remedy is to treat each ARC-AGI task as a \emph{task family}: a distribution over tasks that share the same latent rule while varying nuisance details such as colors, object placement, and grid size.
Sampling fresh tasks from the same family enables matched-distribution benchmarking, robustness sweeps, and dataset-centric analyses that are impossible with a single fixed instance.
Prior work moves in this direction, but important gaps remain.
Augmentation expands ARC-style data via generic symmetries and transformations \cite{augarc2024}, while external/LLM synthesis can scale pretraining data but may drift from the target distribution and does not reliably guarantee that the \emph{set of training examples} disambiguates the intended rule \cite{gifarc2025,barc}.
Programmatic generator frameworks and DSL-based reconstructions provide procedural task families \cite{arcdsl2024,rearc2024}, and concurrent work such as Google’s ARC-GEN pursues broad coverage via mimetic procedural generation \cite{moffitt2025}.
However, existing generator efforts often focus on producing many instances without (a) an explicit, first-class mechanism to enforce \emph{cross-example} constraints that ensure solvable tasks, and (b) solver-facing \emph{step-by-step natural-language reasoning} that remains aligned to each sampled instance.

We introduce \textbf{ARC-TGI} (ARC Task Generators Inventory)\footnote{https://github.com/Omega-Reasoning/ARC-Task-Generators-Inventory} 
, a framework for authoring \emph{human-validated task-family generators} that make task construction and solver-facing supervision explicit.

Each generator is a compact Python module with three stages: (i) sample inputs (randomizing nuisance factors), (ii) apply a deterministic transformation (the latent rule), and (iii) construct the episode by assembling train/test pairs under \emph{task-level constraints}.
Stage (iii) is central: ARC tasks are designed \emph{sets} of examples, not independent input--output pairs.
Accordingly, ARC-TGI enforces constraints that prevent test-only features, reject degenerate shortcuts (e.g., identity or constant outputs), and ensure training pairs collectively expose the variations needed to infer the rule.
In addition, each sampled task is paired with (a) an \emph{input reasoning chain} describing the input grid contents and (b) a \emph{transformation reasoning chain} describing the solution steps.
Both are written as natural-language templates with variable slots that are instantiated from the sampled task parameters, yielding solver-facing explanations aligned to each generated instance.
ARC-TGI also emits partially evaluated Python programs (with task variables inlined) for input sampling, transformation, and task construction, providing a clean interface for code-based solvers or code-conditioned models.

ARC-TGI is paired with a human-in-the-loop creation process: contributors analyze tasks, author reasoning templates and invariants, optionally use an LLM to draft generator code, and iteratively refine generators under repeated sampling and visualization until grids and reasoning traces remain correct and natural under variation.
This \emph{human refinement and validation} is especially important for task-level constraints, where subtle bugs can yield ambiguous or misleading tasks.
Alongside the framework, we release \textbf{461} generators covering \textbf{180} ARC-Mini tasks, \textbf{215} ARC-AGI-1 tasks (200 train, 15 eval), and \textbf{66} ARC-AGI-2 tasks (55 train, 11 eval).
Because each generator defines a distribution rather than a single fixed task, ARC-TGI supports scalable sampling for training or evaluation and enables dataset-centric analyses of \emph{diversity} and \emph{distributional coverage}, with tools that summarize induced variation (e.g., grid sizes, object statistics, and diversity/uniqueness measures).

\section{Related Work}
\label{sec:related_work}

\textbf{ARC and external benchmark variants.}
Several \emph{external} efforts have proposed ARC benchmark variants that largely retain the underlying \emph{static} task set while modifying evaluation structure or task organization.
ConceptARC \cite{conceptarc2023} groups tasks into concept categories for more diagnostic evaluation.
Mini-ARC \cite{miniarc2023} simplifies tasks by reducing grid size (5$\times$5) and adds tooling for human interaction traces, at the cost of reduced coverage.
MC-LARC \cite{mclarc2024} reformulates ARC into multiple-choice questions, changing the original open-ended setting and introducing new shortcut pathways.
ARC-TGI preserves the open-ended format but makes tasks resampleable by representing them as \emph{task families}.

\textbf{Synthetic/augmented ARC-style datasets.}
A common response to ARC’s small size is to generate additional training data.
AugARC \cite{augarc2024} applies systematic symmetries and related transformations to ARC training tasks for large-scale augmentation, but primarily varies surface form and lacks mechanisms for \emph{episode-level} constraints across train pairs (e.g., ensuring the set collectively disambiguates the rule).
GIFARC synthesizes ARC-style tasks from analogy-rich GIFs and pairs them with explicit analogies, but is not designed to faithfully match the distribution of existing ARC(-AGI) tasks \cite{gifarc2025}.
LLM-generated corpora (e.g., ARC-Heavy/BARC) further expand data but may drift in distribution and typically do not provide solver-facing, per-episode reasoning aligned to each sampled instance \cite{barc}.
ARC-TGI provides solver-facing natural-language reasoning templates that are instantiated for each sampled instance.

\textbf{Programmatic generators and DSLs.}
Several projects provide programmatic mechanisms to generate many instances per ARC task.
ARC-DSL introduces a domain-specific language and per-task generators for the 400 public ARC training tasks \cite{arcdsl2024}, and reverse-engineering efforts such as ReARC similarly provide procedural generators for the training set \cite{rearc2024}.
Concurrent with our work, Google released \emph{ARC-GEN}, a mimetic procedural generator aiming for exhaustive coverage of ARC-AGI-1 training tasks while matching the original distribution \cite{moffitt2025}.
ARC-TGI differs in ways that target \emph{human-solvable} episodes: (i) step-by-step natural-language \emph{input} and \emph{transformation} reasoning templates instantiated per sampled instance, and (ii) a first-class \texttt{create\_grids} stage that enforces \emph{episode-level} constraints (e.g., enforcing rule disambiguation and preventing test-only features)
 rather than sampling pairs independently.
ARC-TGI uses human-in-the-loop refinement to validate generators and reasoning traces.

\textbf{Interactive environments, human data, and solver progress.}
ARCLE reformulates ARC as an interactive reinforcement-learning environment, enabling research on action-based solvers and curriculum design, but does not itself define a task-family distribution \cite{arcle2024}.
Human behavioral datasets such as H-ARC provide large-scale traces of human attempts on ARC tasks and are complementary to ARC-TGI’s goal of producing resampled, reasoning-aligned episodes \cite{legris2024harc}.
On the solver side, recent ARC Prize reports and winning approaches highlight continued progress via neurally guided program synthesis and test-time adaptation \cite{arcagi2024,arcprize2024leaderboard,omniarc,lpn2024,ngpt2024,ouellette2024neurallyguided}.
ARC-TGI is intended as evaluation and data infrastructure for this ecosystem.

Overall, a gap remains in jointly supporting task resampling, and episode-level clarity.
Addressing it enables controlled robustness studies and scalable evaluation.

\section{ARC-TGI Framework}
\label{sec:framework}

ARC-TGI operationalizes ARC(-AGI) evaluation in terms of \emph{task families}: parameterized distributions over ARC-style \emph{episodes}.
An episode consists of multiple training and test input/output grid pairs that share the same latent rule but vary in nuisance details such as colors, object placement, and grid size.
The framework design has three goals: (i) \textbf{scalability} (sample many episodes per family), (ii) \textbf{episode solvability} (training examples disambiguate the rule and avoid test-only cues), and (iii) \textbf{interpretability} (solver-facing reasoning aligned to each instance).

\medskip
\noindent\textbf{Generator interface and task-family representation.}
Each ARC-TGI task family is implemented as a compact Python module (single file) that subclasses an abstract \texttt{ARCTaskGenerator} and implements three core methods with fixed signatures.
To improve readability, we summarize the interface in Table~\ref{tab:api}.

\begin{table}[h!]
\centering
\small
\setlength{\tabcolsep}{6pt}
\begin{tabular}{p{0.25\linewidth} p{0.60\linewidth}}
\toprule
\textbf{Method} & \textbf{Role (fixed signature)} \\
\midrule
\texttt{create\_input} &
\texttt{(self, taskvars: dict, gridvars: dict) -> np.ndarray}: sample an input grid while randomizing nuisance factors. \\
\texttt{transform\_input} &
\texttt{(self, grid: np.ndarray, taskvars: dict) -> np.ndarray}: deterministically apply the latent rule to produce the output grid. \\
\texttt{create\_grids} &
\texttt{(self) -> Tuple[Dict[str, Any], TrainTestData]}: construct a complete episode (train/test pairs) and return sampled variables plus train/test data. \\
\bottomrule
\end{tabular}
\caption{ARC-TGI generator API. \texttt{taskvars} are per-episode variables used to instantiate templates; \texttt{gridvars} are per-grid auxiliaries used during sampling.}
\label{tab:api}
\end{table}

This separation is deliberate.
\texttt{create\_input} may use rich randomness to increase diversity, whereas \texttt{transform\_input} is a concise, solver-facing transformation that should not depend on the randomness of generation-time.
The \texttt{create\_grids} stage then assembles the episode (training and test pairs) and is the right place to enforce constraints that couple examples across the episode.

\medskip
\noindent\textbf{From a generator to a concrete task instance.}
A concrete episode is produced by calling a base-class wrapper (e.g., \texttt{create\_task}) around \texttt{create\_grids()}.
The framework emits a standardized record containing:
(i) an instantiated \emph{input reasoning chain},
(ii) an instantiated \emph{transformation reasoning chain},
(iii) partially evaluated Python programs for input sampling, transformation, and episode construction (with sampled variables inlined), and
(iv) the resulting train/test grids.
Repeated calls to \texttt{create\_task} sample different instances from the same task family.

\medskip
\noindent\textbf{Reasoning templates aligned to sampled instances.}
Beyond grids, ARC-TGI attaches solver-facing natural-language explanations to each sampled episode.
Each generator defines two lists of strings: an \textbf{input reasoning chain} (``what is in the input grid?'') and a \textbf{transformation reasoning chain} (``how do we get the output?'').
These strings can include template expressions instantiated from \texttt{taskvars} (e.g., color names, grid/object sizes, counts, or relative positions), yielding reasoning traces that remain consistent under nuisance variation.
This supports step-by-step, human-readable descriptions that (a) remain stable under resampling and (b) can be used as supervision or diagnostics when models fail.
ARC-TGI also exports solver-facing code via partial evaluation: parameters sampled in \texttt{create\_grids} are in-lined so the emitted program depends only on the input grid at runtime.
This yields a clean interface for program-based or code-conditioned solvers while preserving randomized sampling across instances.

\medskip
\noindent\textbf{Episode construction and task-level constraints (\texttt{create\_grids}).}
A key design choice in ARC-TGI is to elevate \texttt{create\_grids} to a first-class stage.
ARC(-AGI) episodes are not simply sets of independent input--output pairs: the \emph{collection} of training examples must jointly disambiguate the rule.
Independent pair sampling can yield ambiguous episodes (e.g., missing critical variation, unseen test features, or degenerate shortcuts), so ARC-TGI lets generators enforce \emph{episode-level constraints} across train/test pairs, including:
\begin{itemize}[leftmargin=*]
  \item \textbf{Train--test consistency:} forbid test-only colors/shapes or other cues that make the episode unsolvable from the given training examples;
  \item \textbf{Input construction constraints:} generate inputs under explicit constraints so that the transformation yields a valid and interpretable output;
  \item \textbf{Disambiguating coverage:} ensure training pairs collectively include the variations required to infer the rule;

\end{itemize}
Operationally, \texttt{create\_grids} resamples until constraints are met (rejection sampling), and the framework provides utilities to support this pattern. These are semantic, generator-specific conditions for solvability; separately, ARC-TGI supports framework-level verification of exported episodes (below).

\medskip 
\noindent\textbf{Reusable libraries for generation and transformation.}
ARC-TGI provides two optional helper libraries to promote consistency.
The \textbf{input} library supports diverse, and valid grid sampling (e.g., retry/rejection helpers, contiguous object synthesis under 4-/8-connectivity, extent constraints, controlled coloring/densities) and is used in \texttt{create\_input} but \emph{not} in \texttt{transform\_input}.
The \textbf{transformation} library provides solver-facing, object-centric primitives (e.g., connected components via \texttt{find\_\allowbreak connected\_\allowbreak objects}, \texttt{GridObject}\allowbreak/\texttt{GridObjects}, and geometric/compositional ops such as translation, rotation, reflection, bounding boxes, border behavior, filtering) and can be used both for constraint checking and for concise transformations.

\begin{figure*}
  \centering
    \includegraphics[width=\linewidth]{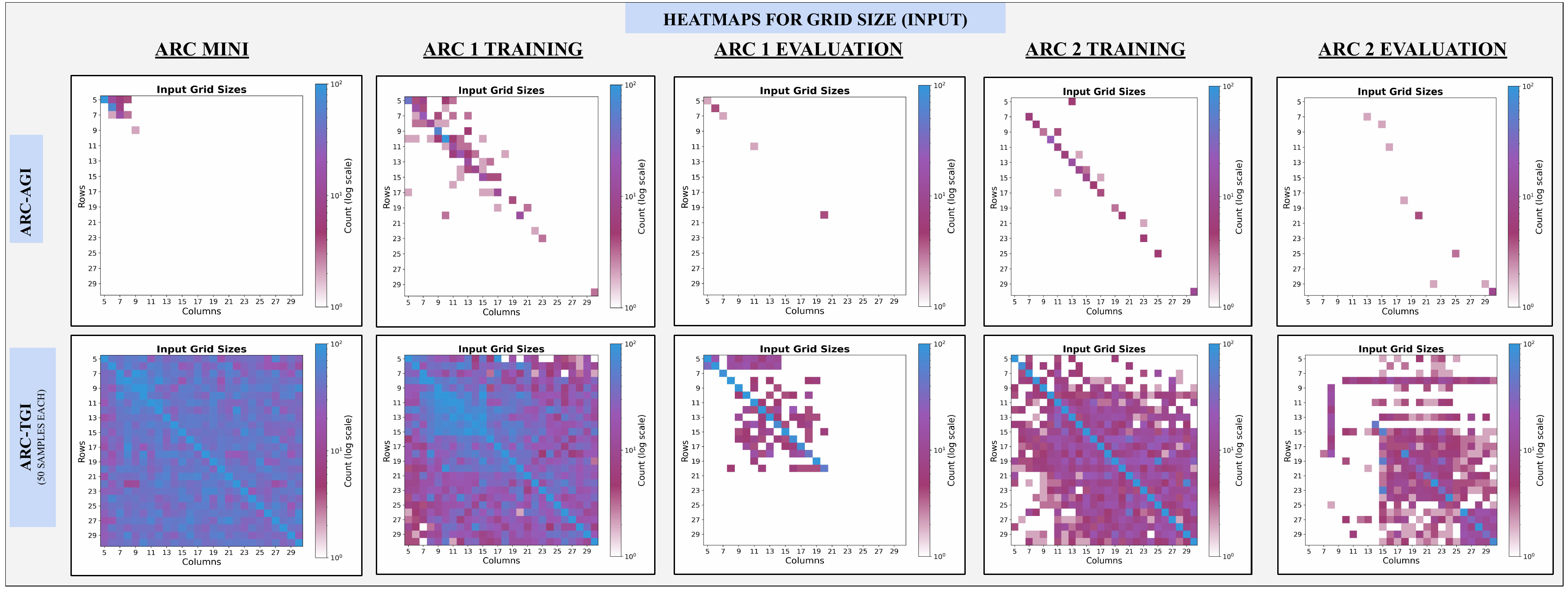}
    \caption{Input grid-size heatmaps. Upper row: original tasks. Lower row: ARC-TGI samples (50 per generator).}
  \label{fig:arcgen_heatmaps_inputs}
\end{figure*}

\medskip\noindent\textbf{Self-verifying exports.}
ARC-TGI exports each sampled episode together with an \emph{executable witness}: an inlined transformation program (plus instantiated reasoning traces) corresponding to the sampled variables.
This makes it possible to run inexpensive, framework-level \emph{verification} during generator development and large-scale sampling, without relying on manual inspection:
\begin{itemize}[leftmargin=*]
    \item \textbf{Executable witness check:} re-running the inlined program on each input grid must exactly reproduce the stored outputs (train and test).
    \item \textbf{Invariant checks:} generic structural invariants (e.g., well-formed grids; declared train/test restrictions such as ``no unseen colors at test'') are automatically verified.
    \item \textbf{Shortcut screening:} we optionally screen for unintended triviality (e.g., identity/constant outputs) and keep it only for families where it is the intended rule.
\end{itemize}

\noindent\textbf{Human-in-the-loop authoring and validation.}
ARC-TGI uses a human-in-the-loop creation process.
Contributors begin with task analysis: they write input/transformation reasoning chains, identify task and grid variables, and specify episode-level constraints.
An LLM may draft generator code, but each generator is iteratively refined and validated through repeated sampling and visualization
until both grids and reasoning traces remain correct and natural under variation.
This workflow targets two common failure modes in procedural generation: subtle rule violations under edge-case randomization and reasoning traces that drift from the sampled instance.
Overall, ARC-TGI combines episode-level constraints with human validation to preserve the defining ARC property of \emph{human-solvable tasks inferred from the training examples}.

\section{Dataset}
\label{sec:analysis}
\medskip\noindent\textbf{Release.} We release a generator suite of 461 task-family generators (one Python file per family) covering ARC-Mini (train), ARC-AGI-1 (train/eval), and ARC-AGI-2 (train/eval), plus shared input-sampling utilities and solver-facing transformation primitives. Each generator can be resampled to produce fresh ARC-JSON tasks and can optionally export (i) instantiated natural-language reasoning traces and (ii) partially evaluated Python programs with sampled task variables inlined.

\medskip\noindent\textbf{License.} We release ARC-TGI under the Apache-2.0 license, consistent with upstream ARC-AGI releases.

\medskip\noindent\textbf{Human-in-the-loop development.} Generators (and their reasoning templates) were iteratively refined by human authors over the past 14 months by repeatedly resampling many random instances and fixing both grid validity and explanation faithfulness under variation. In 40\% of cases, LLMs such as GPT-5.x and Opus 4.5 produced working code on the first try; in 65\% of cases, after multiple prompt iterations; and in 35\% of cases, the generator code was partially manually written. 

\medskip\noindent\textbf{Sample analysis.} For the analysis in this section, 
we sample \(k=50\) tasks per generator, yielding \(461 \times 50 = 23{,}050\) sampled tasks.
We focus on three questions: \textbf{validity} (are samples internally consistent and non-degenerate),
\textbf{coverage} (how sampled grids compare to the original ARC(-AGI) distributions), and
\textbf{diversity} (how much variation appears within a task family).

\medskip\noindent\textbf{Distributional summaries and coverage.}
We summarize global grid-size statistics and compare sampled tasks to the original benchmarks (restricted to families for which a generator exists).
\Cref{fig:arcgen_heatmaps_inputs} shows input grid-size heatmaps where the upper panels aggregate the original (human-authored) examples and the lower panels aggregate ARC-TGI samples (50 draws per generator).
To construct the heatmaps, we aggregate at the \emph{example level}: for every task, we record the row/column dimensions for every input and output grid across all training and test pairs.
This visualization highlights that original benchmarks concentrate on a sparse set of dimensions, while ARC-TGI preserves dominant modes (notably the diagonal where input and output sizes match) and expands coverage by interpolating through controlled within-family variation.

For comparability across splits and plots, we restrict the visualization to grids in the \(5 \times 5\) to \(30 \times 30\) range, which captures the vast majority of examples in both the original benchmarks and ARC-TGI samples; extreme sizes outside this range are rare and do not materially affect the qualitative conclusions.

\begin{figure}[t]
\centering
\includegraphics[width=\columnwidth]{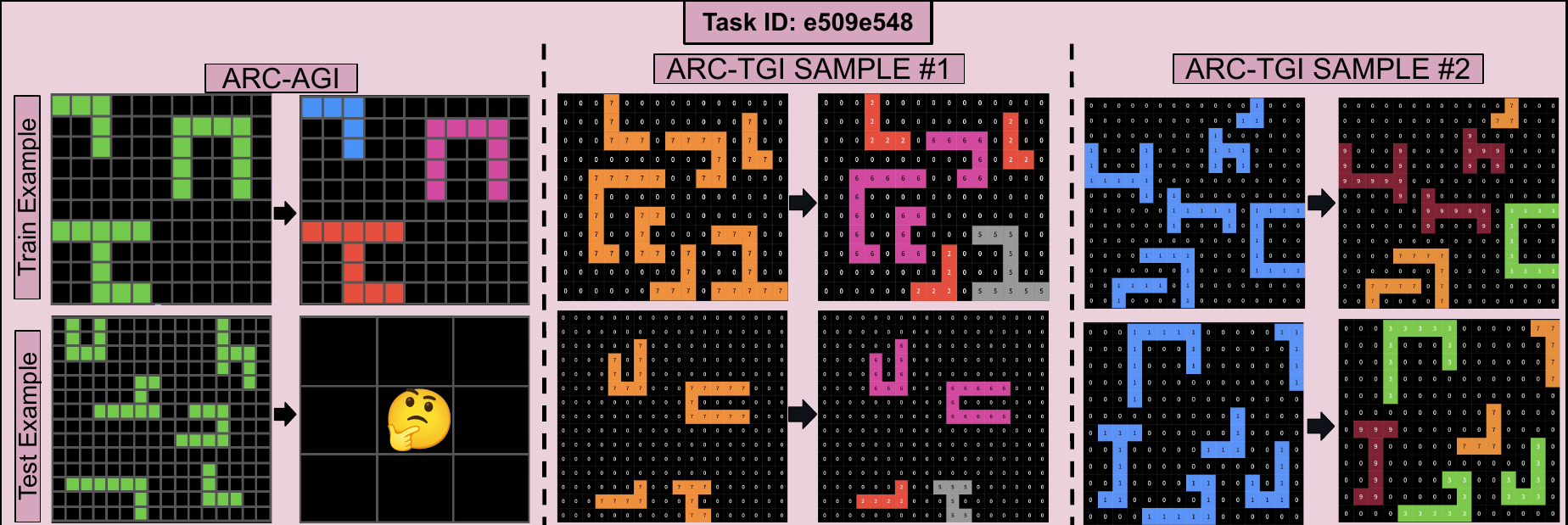}
\caption{ARC-AGI task (Task ID: e509e548) and corresponding ARC-TGI sample.}

\label{fig:within_family_variation}
\end{figure}

\begin{figure}[t]
    \centering
    \includegraphics[width=\columnwidth]{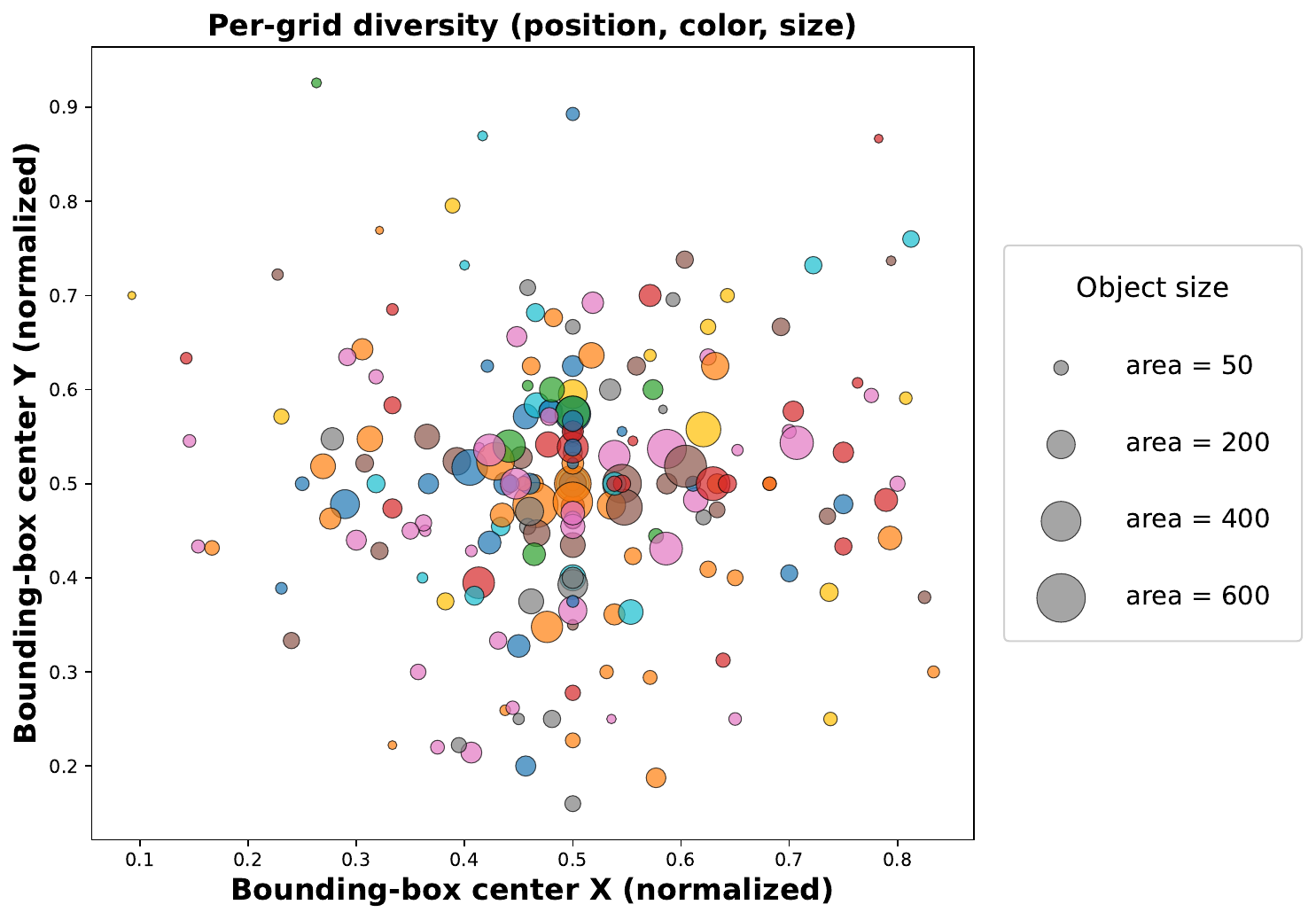}
    \caption{Per-grid diversity visualization for one ARC-TGI generator (\texttt{taskcmBhVbGzL8ZgWXDE5CUS6B}), showing variation in spatial position, size, and color across sampled grids while preserving the latent rule.}
    \label{fig:td_plot}
\end{figure}

\medskip\noindent\textbf{Diversity within task families.}
To illustrate within-family variation induced by a single generator, we visualize grids sampled from one ARC-TGI family (task ID: \texttt{taskcmBhVbGzL8ZgWXDE5CUS6B}).
Rather than collapsing each task to a single averaged summary, we represent \emph{each individual input grid} (from both training and test examples) as a point in a simple feature space capturing: (i) spatial position (bounding-box center), (ii) scale (foreground area or occupied pixels), and (iii) palette (dominant object color).
\Cref{fig:td_plot} shows that a single family can vary placement, size, and colors substantially while preserving the same latent transformation rule (as evidenced by the consistent input--output relationship within each sampled task).
This breadth is important for benchmarking: solvers must learn the abstract rule rather than relying on fixed spatial or color-specific heuristics.

\section{Experiments}
\label{sec:experiments}
We conducted two complementary studies in order to validate the quality and utility of our generators: 
(1) evaluation on the generated dataset of pre-trained Decoder LLMs without fine-tuning, and
(2) assessment of the impact of fine-tuning with the ARC-TGI dataset, 

\noindent\textbf{Training Configuration.} 
A consistent training recipe was used for all fine-tuning experiments. 
Parameter-efficient fine-tuning using LoRA was employed with rank r = 64, $\alpha$ = 16, and dropout = 0.05. 
The LLMs were trained for 10 epochs with double quantization, using the AdamW optimizer with a learning rate of 2e-5, a warmup ratio of 0.03, and zero weight decay. A maximum context length of 14,000 tokens, along with 2048 tokens for generation length, was used to constrain the models. 
Moreover, for fine-tuning and evaluation across all three studies, we used few-shot prompting with 3-5 training examples. 
Model input, prompt setting, and evaluation method are reported in Fig.\ref{fig:es1}. 
Finally, all experiments were conducted on H100 and H200 GPUs.

\noindent\textbf{Choosing the per-generator sampling budget.}
\label{sec:sampling_budget}
We performed a small sweep over the number of sampled tasks per generator
(\(n \in \{20,40,50,70,100,150\}\)) to select a reasonable default for subsequent experiments.
In our current runs, \(n=50\) yielded the strongest improvement, while other values were close to the baseline;
we therefore use \(n=50\) as a practical setting rather than a statement about scaling behavior.
A more systematic study of scaling with multiple random seeds and controlled filtering is left for future work.

\subsection{Model performance on the generator dataset}
\begin{figure}[t!]
    \centering
    \includegraphics[width=\columnwidth]{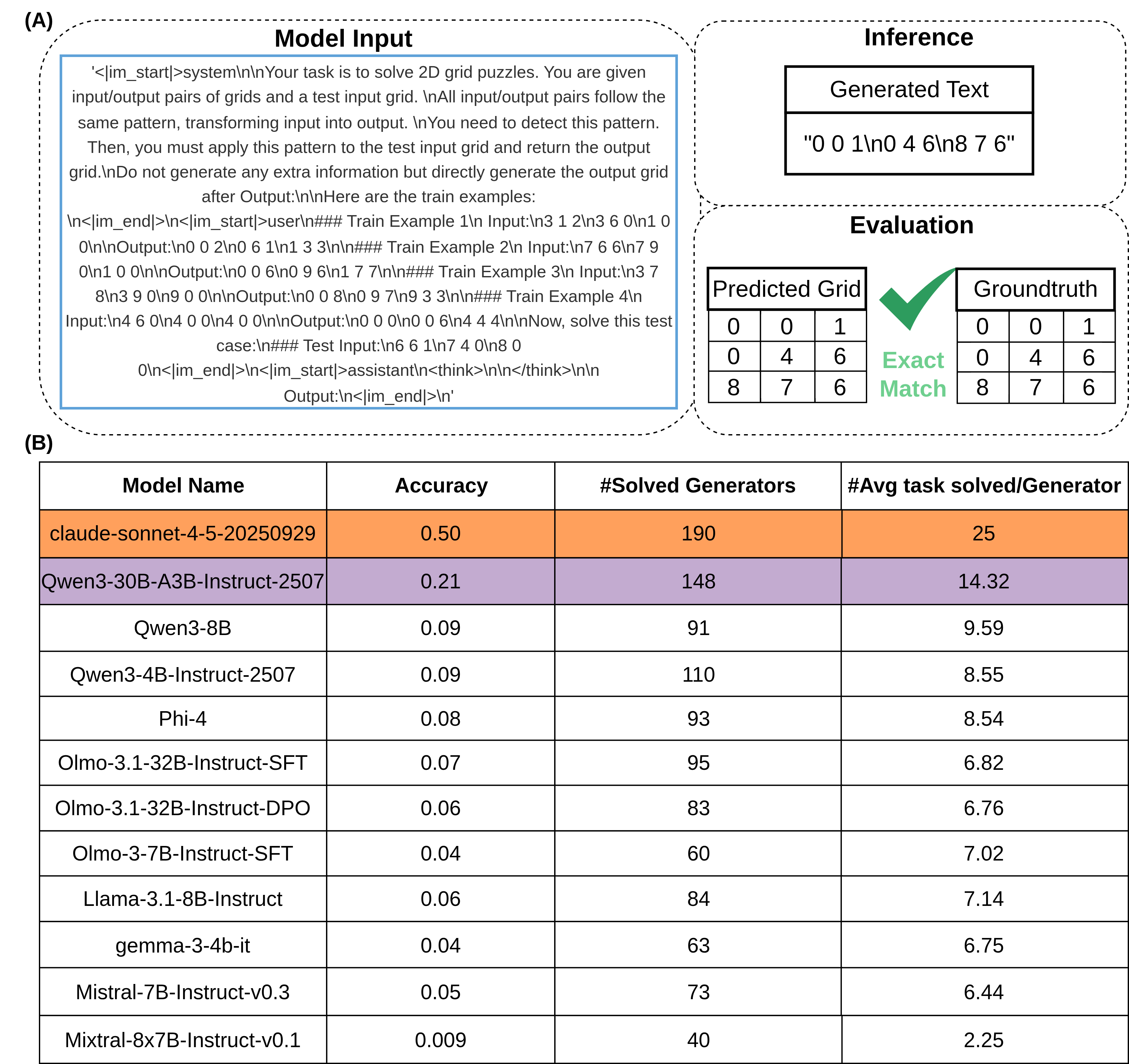}
    \caption{\textbf{Few-shot performance on \textsc{ARC-TGI-50N}.} (A) ARC-style episode: infer a latent rule from training input/output pairs and apply it to a test input. (B) Exact-match accuracy across models on \textsc{ARC-TGI-50N}.}
    \label{fig:es1}
\end{figure}

We evaluate how pre-trained LLMs perform on ARC-style 2D grid puzzles under few-shot prompting.
Given the training input/output pairs, models must infer the latent rule and apply it to the test input to produce the output (Fig.~\ref{fig:es1}).
For evaluation, we sample 50 episodes from each of 200 generators (10{,}000 total), denoted \textsc{ARC-TGI-50N}, and filter instances that exceed a 14{,}000-token context limit leading to approximately 400 filtered samples. 
During benchmarking, solvers are given only the ARC-style grids (training I/O pairs and the test input), i.e., reasoning traces are not provided at inference time.

\noindent\textbf{Experimental Setup.} 
We evaluated 11 open-source decoder LLMs (4B--32B parameters) on datasets sampled from our generator collection across multiple task families.
Models were chosen for strong reported reasoning performance on established math and reasoning benchmarks and span recent families including Qwen-3, Gemma-3, Mistral/Mixtral, Llama-3, Phi-4, and Olmo-3.
Specifically, we evaluate:
(1) Qwen3-4B-Instruct-2507 \cite{qwen3_2025},
(2) Qwen3-8B \cite{qwen3_2025},
(3) Qwen3-30B-A3B-Instruct-2507 \cite{qwen3_2025},
(4) Olmo-3.1-32B-Instruct-SFT \cite{olmo3_2025},
(5) Olmo-3-7B-Instruct-SFT \cite{olmo3_2025},
(6) Olmo-3.1-32B-Instruct-DPO \cite{olmo3_2025},
(7) Phi-4 \cite{phi4_2024},
(8) Mixtral-8x7B-Instruct-v0.1 \cite{mixtral_2024},
(9) Mistral-7B-Instruct-v0.3 \cite{mistral7b_2023},
(10) Llama-3.1-8B-Instruct \cite{llama3_2024},
and (11) gemma-3-4b-it \cite{gemma3_2025}.
For a closed-source reference point, we also evaluated Claude Sonnet 4.5 on the same ARC-TGI dataset.

\begin{figure}[h!]
    \centering
    \includegraphics[width=\columnwidth]{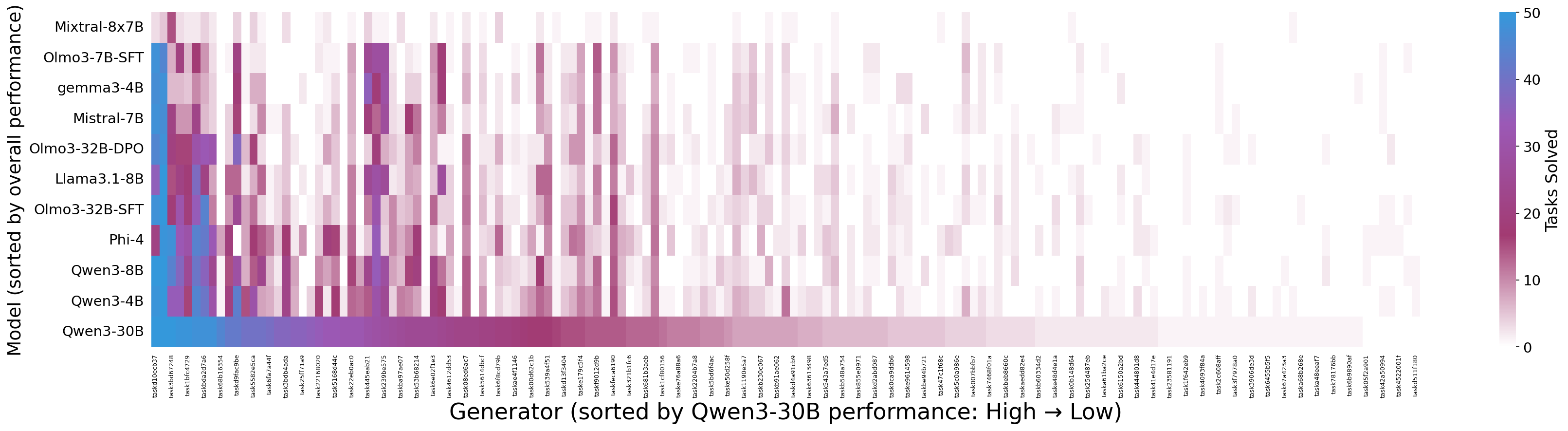}
    \caption{\textbf{Per-generator performance across models (200 generators).} Heatmap of model accuracy by generator. Models are ordered by mean accuracy (y-axis) and generators are ordered by Qwen3-30B accuracy (x-axis), revealing shared easy/hard regions and sparse success on the hardest generators.}
    \label{fig:es2}
\end{figure}

\noindent\textbf{Result Analysis.}
Figure~\ref{fig:es1} reports results on \textsc{ARC-TGI-50N}.
We use \textbf{exact-match accuracy}, i.e., the predicted output grid must match the ground truth grid exactly.
Overall, we observe a pronounced hierarchy both \emph{within} model families by parameter size and \emph{across} model families.
The best open-source performance is achieved by Qwen3-30B (21\%), which opens a clear gap over the remaining models.
Across families, 8B models generally outperform 4B models, indicating that parameter scale matters for these few-shot 2D grid transformations.
The Qwen family performs best across all parameter sizes.
In contrast, Mixtral attains the lowest performance (0.09\% accuracy), and Olmo-32B (both SFT and DPO) is comparatively weak relative to Qwen3-30B despite similar scale.

These results suggest that ARC-TGI’s 2D grid transformations remain challenging for current few-shot LLM prompting.
Also, models solve only a few tasks per generator on average (2.25--14.32; Fig.~\ref{fig:es1}), and only a few generators yield high task accuracy.

\noindent\textbf{Task-family coverage and task-level structure.}
Figure~\ref{fig:es2} provides task-level insight via a generator-by-model heatmap.
Models (y-axis) are ordered by average performance, while generators (x-axis) are sorted by Qwen3-30B performance.
Two patterns stand out.
First, Qwen3-30B achieves the broadest coverage, solving tasks from 148/200 generators, yet high accuracy is concentrated in a relatively small subset of generators.
Second, generator difficulty is strongly shared across models: the easiest generators (solved by almost all models) cluster on the left (deep blue), a middle band is partially solved by many models (pinkish region), and the hardest generators appear on the far right (nearly white), solved by few or no models.
Across all models, between 40--148 generators are solved, indicating that the generator suite induces diverse transformations that different model architectures recognize differently; nevertheless, most models tend to succeed on largely overlapping subsets of generators, with substantial variation in consistency (rate of successful tasks per generator).
Mixtral is the least consistent, with the lowest accuracy (below 1\%) and a dramatically lower frequency of solved tasks per generator.

\begin{figure}[t]
    \centering
    \includegraphics[width=\columnwidth]{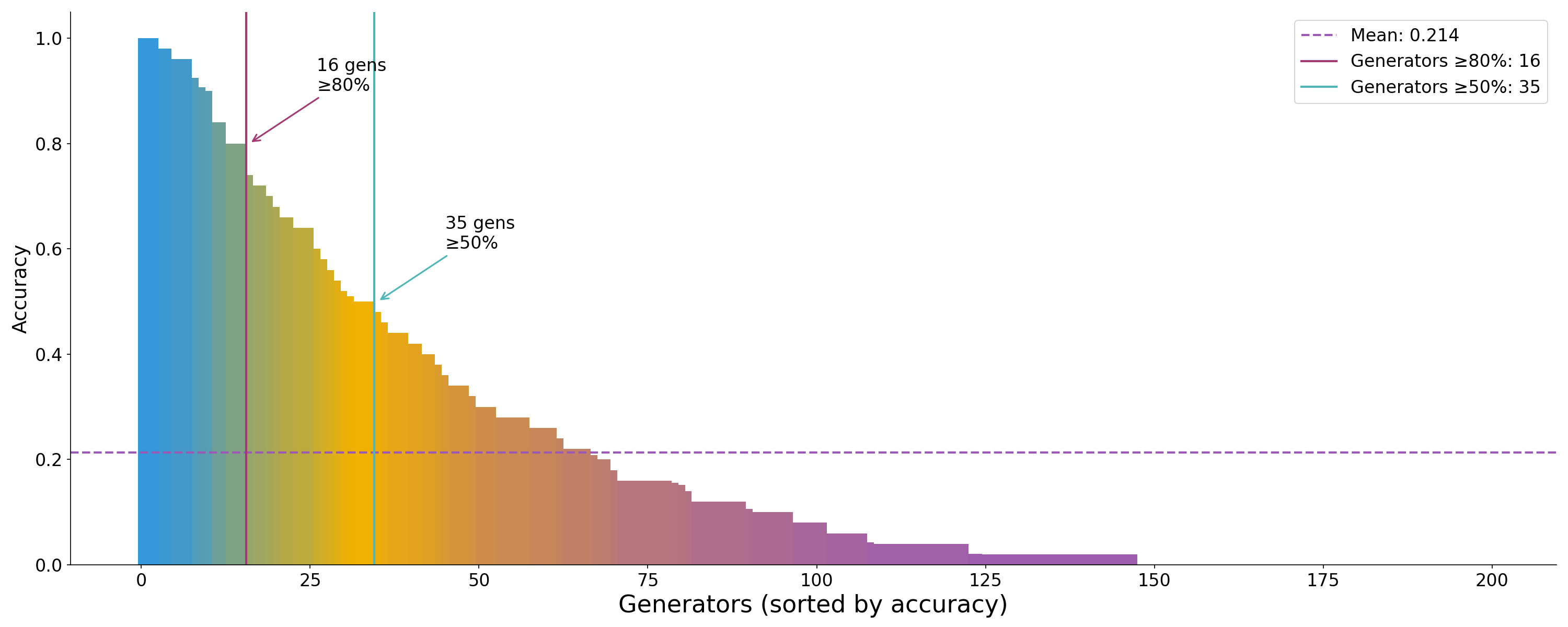}
    \caption{\textbf{Per-generator accuracy for Qwen3-30B.} Accuracy over generators (50 samples per generator), highlighting broad coverage with a long-tail of difficult task families.}
    \label{fig:es3}
\end{figure}

\noindent\textbf{Long-tail difficulty under resampling (Qwen3-30B).}
Figure~\ref{fig:es3} reports per-generator accuracy for Qwen3-30B and shows substantial variance across generators.
The model solves at least one task for 148/200 generators, but performance has a long-tail: among these 148 generators, accuracy exceeds 80\% for only 16 generators and exceeds 50\% for 35 generators.
Among the 52 generators with zero solved tasks, 49 are also unsolved in the original ARC-AGI-1 training set, suggesting that generator-specific difficulty is preserved under resampling.
Overall, Qwen3-30B engages with many task families, but its efficiency varies sharply across them.

\begin{figure}[t]
    \centering
    \includegraphics[width=\columnwidth]{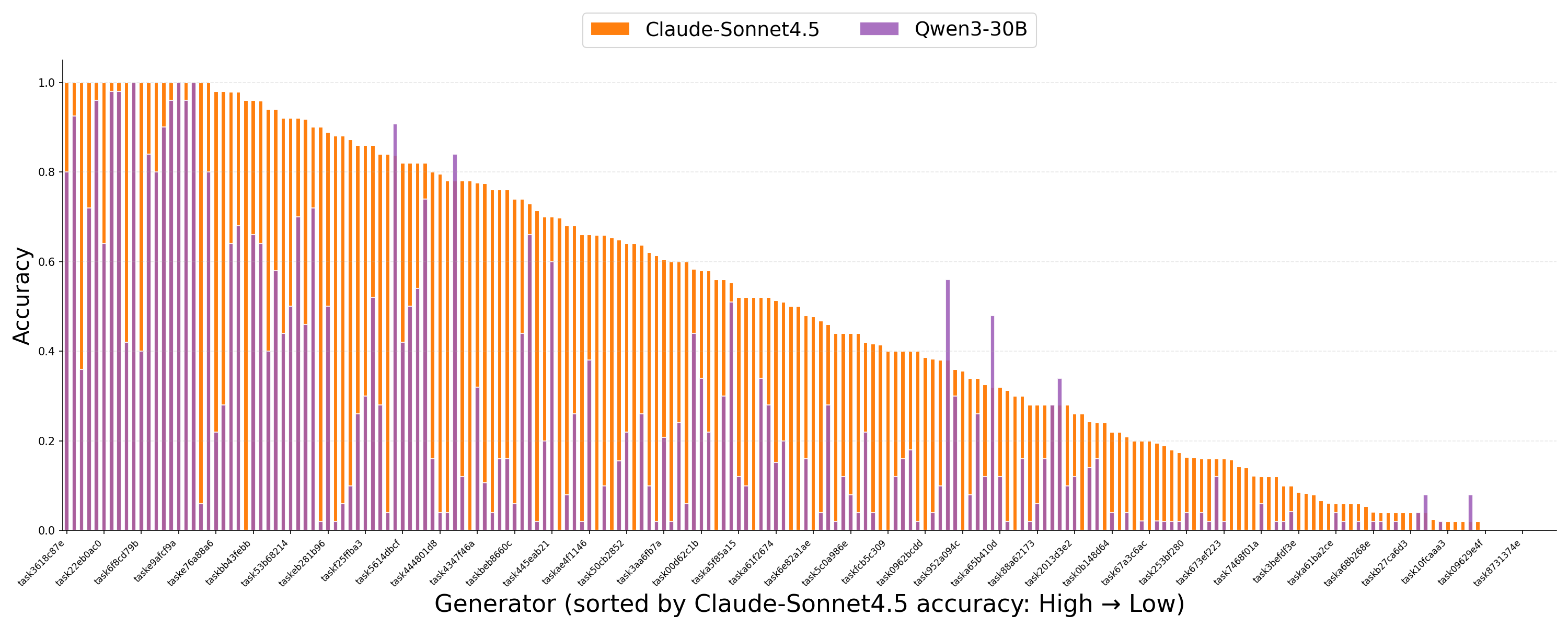}
    \caption{\textbf{Per-generator comparison: Qwen3-30B vs.\ Claude Sonnet 4.5.} Each point summarizes accuracy on a generator (50 samples), showing largely consistent generator difficulty with occasional reversals.}
    \label{fig:es4}
\end{figure}
\begin{figure}[t]
    \centering
    \includegraphics[width=\columnwidth]{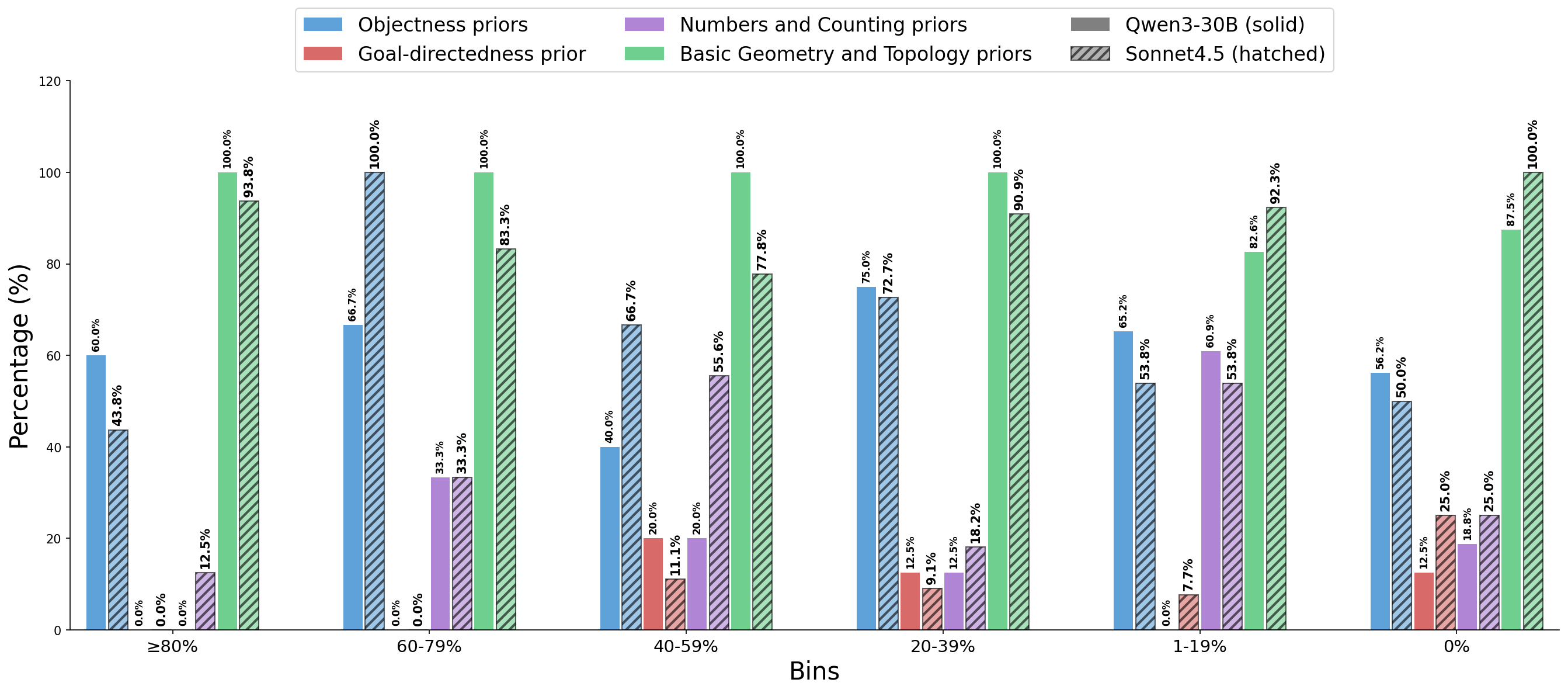}
    \caption{\textbf{Core-knowledge priors vs.\ accuracy.} Distribution of dominant priors across accuracy bins for Qwen3-30B and Sonnet 4.5 (manual annotation on stratified samples).}
    \label{fig:prior_comparison_SonnetVsQwen}
\end{figure}

\begin{figure*}[t]
    \centering
    \includegraphics[width=\textwidth]{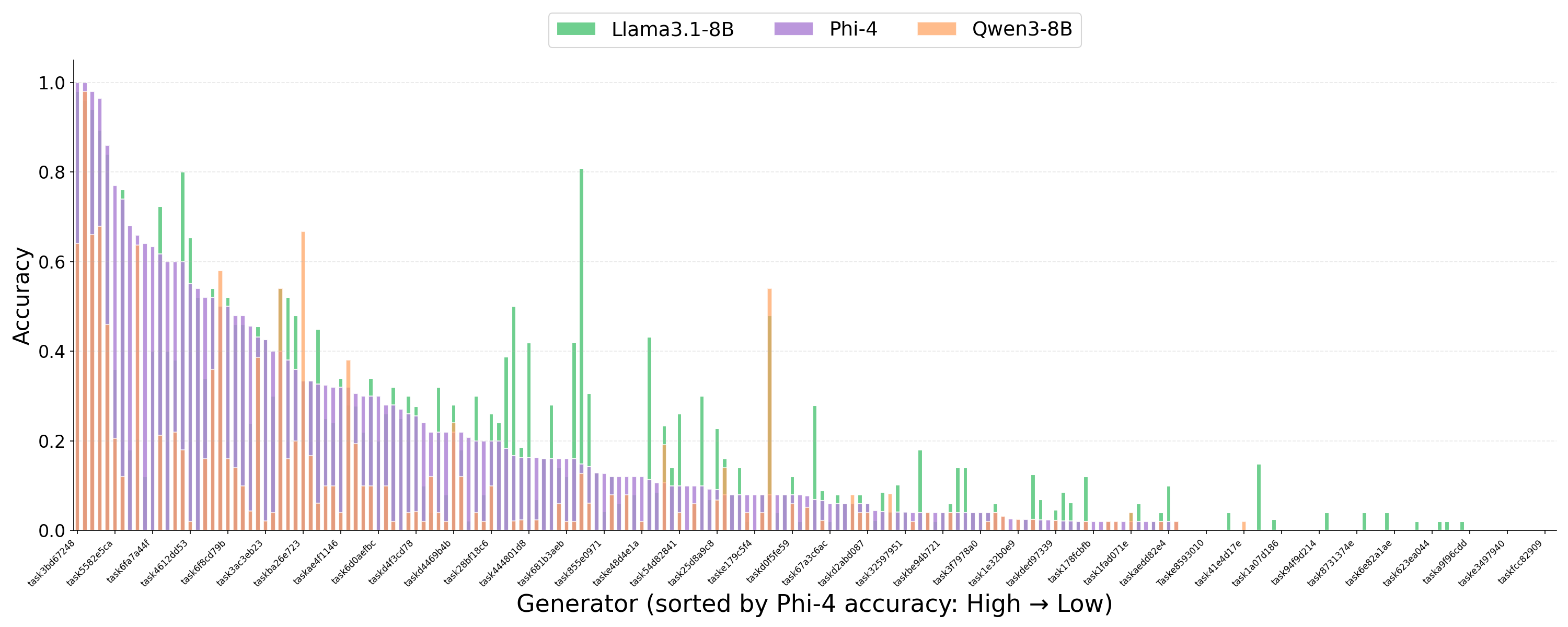}
    \caption{\textbf{Post-fine-tuning task coverage across 200 generators.} Stacked bars summarize task-level accuracy across generators after fine-tuning on \textsc{ARC-TGI-50N}.}
    \label{fig:s2f3}
\end{figure*}

\begin{figure}[t]
    \centering
    \includegraphics[width=\columnwidth]{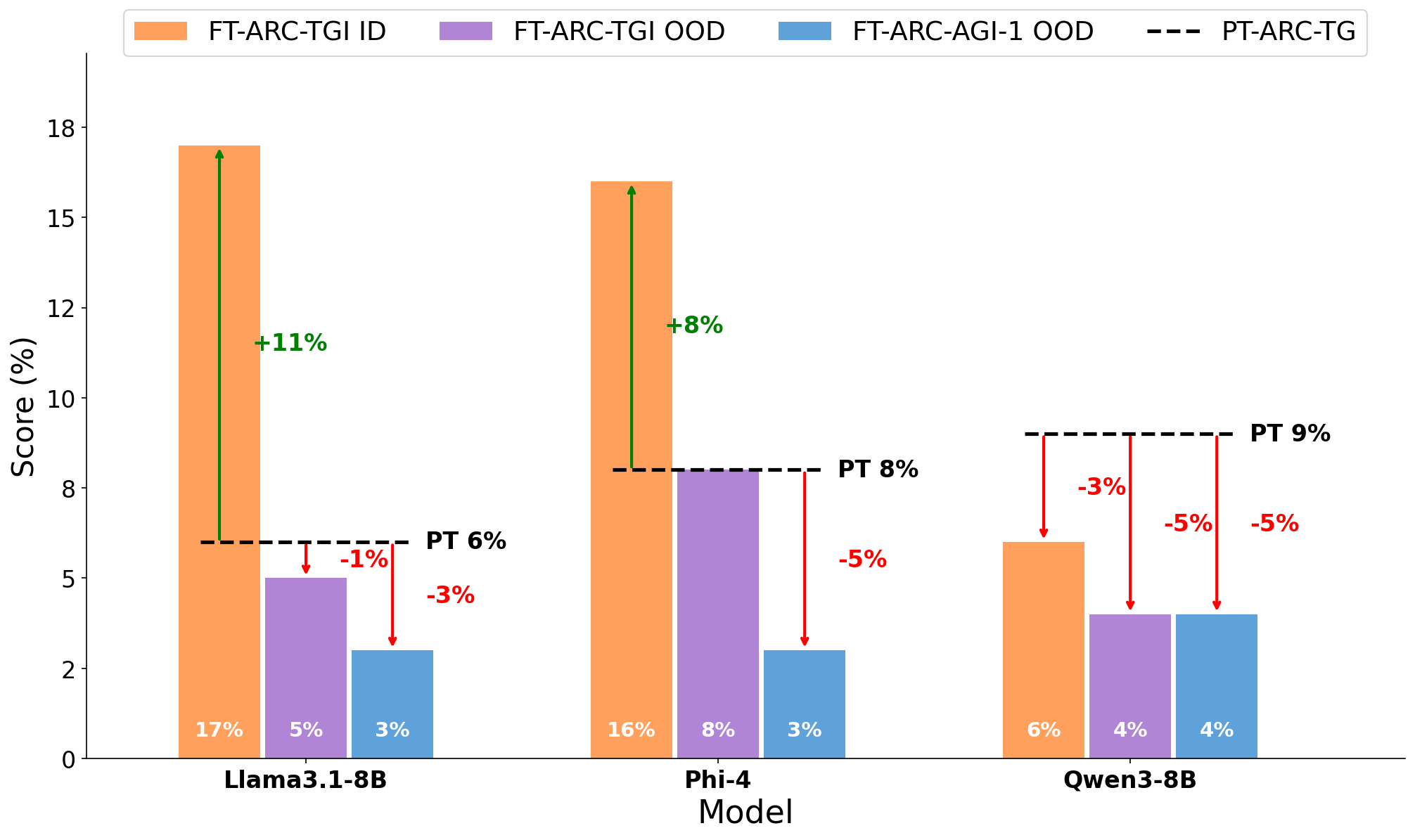}
    \caption{\textbf{Fine-tuning effects on ID vs.\ OOD evaluation.} Accuracy for pre-trained models on ARC-TGI (PT-ARC-TGI), after ARC-TGI fine-tuning evaluated ID (FT-ARC-TGI ID) and on ARC-AGI-1 eval (FT-ARC-TGI OOD), and after fine-tuning on ARC-AGI-1 train evaluated on ARC-AGI-1 eval (FT-ARC-AGI-1 OOD).}
    \label{fig:s2f1}
\end{figure}
\begin{figure}[t!]
    \centering
    \includegraphics[width=\columnwidth]{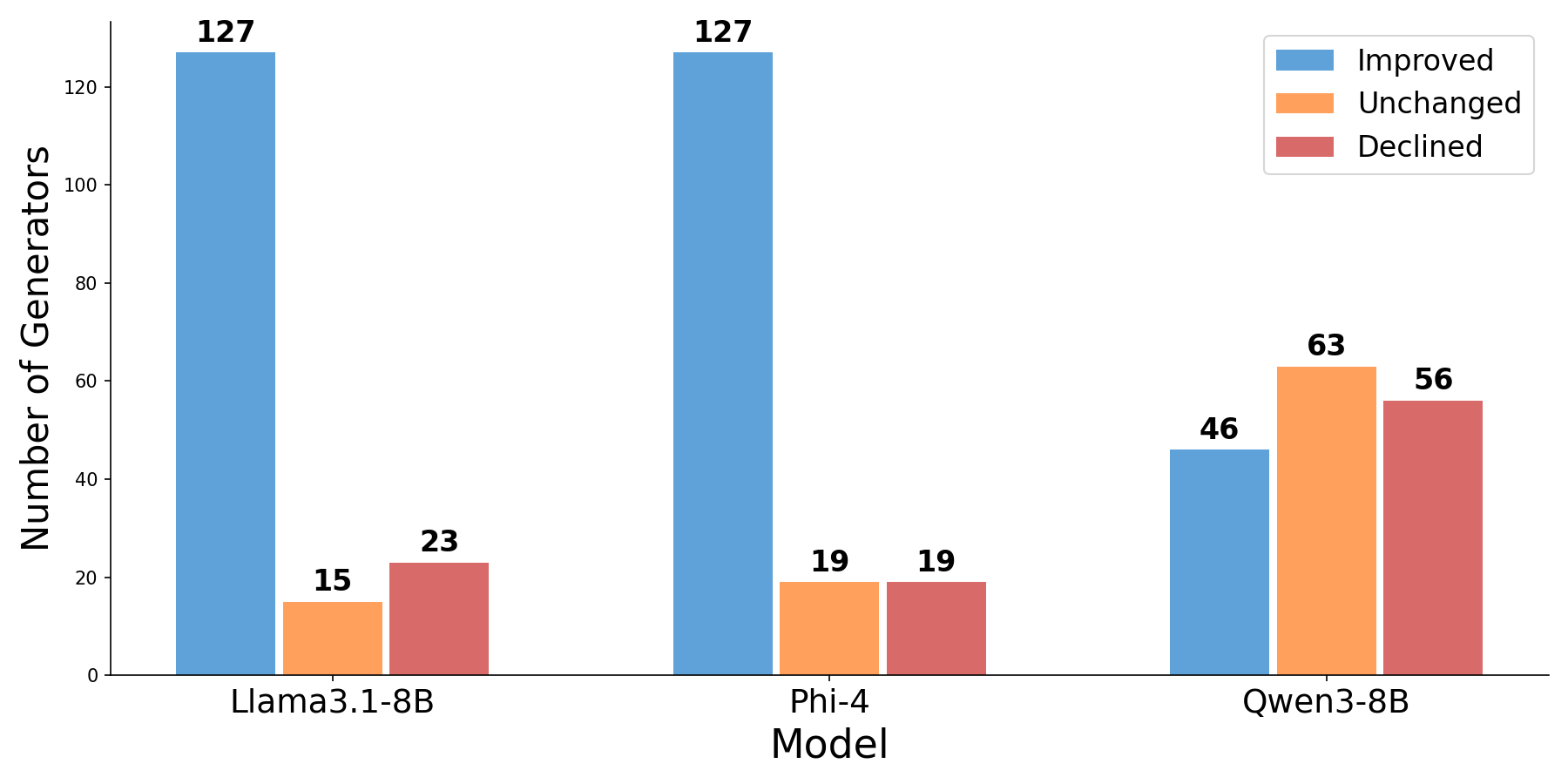}
    \caption{\textbf{Per-generator change after ARC-TGI fine-tuning.} Number of generators whose accuracy improves, declines, or remains unchanged (per model).}
    \label{fig:s2f2}
\end{figure}
\noindent\textbf{Closed-source reference and generator difficulty consistency.}
To benchmark the leading open model (Qwen3-30B), we compare per-generator performance to Claude Sonnet 4.5 in Fig.~\ref{fig:es4}.
Both models induce a similar difficulty ranking over generators, suggesting that task difficulty is generator-specific and consistent under sampling.
Despite being much smaller, Qwen3-30B outperforms Sonnet 4.5 on some generators.
Overall, Sonnet 4.5 solves 190/200 generators and achieves an average solved-task rate of 50\% per generator (Fig.~\ref{fig:es1}), supporting rule consistency and robust sampling in ARC-TGI.
The remaining gaps and low solved-task frequency for many open models reinforce the value of ARC-TGI for probing reasoning limits under controlled resampling.

\noindent\textbf{Core-knowledge priors and difficulty.}
To analyze sources of performance variability, we annotate task difficulty using the core knowledge priors of \cite{chollet2019} (Fig.~\ref{fig:prior_comparison_SonnetVsQwen}).
Both models exhibit qualitatively similar trends across accuracy bins.
Tasks with accuracy above 80\% show little involvement of number/counting priors and generally lack goal-directedness priors; as accuracy decreases, both number/counting and goal-directedness priors become more prevalent.
Lower-accuracy bins more frequently require integrating multiple priors, suggesting that higher compositionality is associated with increased difficulty.
These insights are based on stratified sampling across accuracy levels, followed by manual identification of dominant priors by two authors.

\subsection{Fine-tuning with ARC-TGI}
The goal of this study is to test whether fine-tuning on ARC-TGI improves LLM reasoning on ARC-style 2D grid puzzles, and whether any gains transfer to an external benchmark.

\noindent\textbf{Experimental setup.}
Based on the pre-trained results (and for computational efficiency), we fine-tune three 8-14B-class open models: Qwen3-8B, Llama-3.1-8B, and Phi-4.
We evaluate under two complementary settings:
\begin{enumerate}[leftmargin=*]
    \item \textbf{ID (in-distribution):} from the same 200 generators, we sample 100 tasks per generator and split the resulting dataset 50/50 into train and test.
    \item \textbf{OOD (out-of-distribution):} we evaluate on the public ARC-AGI-1 evaluation set, comparing two fine-tuning regimes:
    (i) \textbf{FT-ARC-TGI} (models fine-tuned on ARC-TGI samples) and
    (ii) \textbf{FT-ARC-AGI-1} (models fine-tuned on the public ARC-AGI-1 training set).
\end{enumerate}

\noindent\textbf{Result analysis.}
Figure~\ref{fig:s2f1} summarizes accuracy across the evaluation setups.
Fine-tuning on ARC-TGI yields large improvements relative to pre-trained performance on ARC-TGI: Phi-4 nearly doubles accuracy (+100\% relative; 8\%$\rightarrow$16\%), and Llama-3.1-8B improves even more (+183\% relative; 6\%$\rightarrow$17\%).
In contrast, Qwen3-8B shows an unexpected degradation of $\approx$33\% (9\%$\rightarrow$6\%), suggesting that architectural differences influence how models benefit from ARC-TGI fine-tuning. 
Comparing ID and OOD settings reveals a consistent gap: ID performance is higher than OOD performance for all three models, with differences ranging from 100\% (Phi-4) to 240\% (Llama-3.1-8B).
Among the three, Phi-4 exhibits the strongest transfer when fine-tuned on ARC-TGI.  Overall, the results indicate that fine-tuning helps models solve more tasks but generalization to new puzzle distributions remains limited.
Finally, comparing fine-tuning sources on the OOD benchmark shows that FT-ARC-TGI yields better generalization for Phi-4 (167\% improvement) and Llama-3.1-8B (67\% improvement), while Qwen3-8B remains unchanged.

\noindent\textbf{Per-generator improvements explain model-level gains.}
Figure~\ref{fig:s2f2} reports how fine-tuning shifts each model's per-generator performance.
Llama-3.1-8B and Phi-4 show a consistent positive pattern: performance improves for 127 generators, declines for 19--23 generators, and remains unchanged for 15--19 generators (respectively).
Thus, fine-tuning increases the ability of both models to solve a larger set of task families.
With Qwen3-8B shows improvement for only 47 generators while declining for 56 (with 63 unchanged)--which directly explains its degraded average accuracy: the model loses capability on more generators than it gains.

\noindent\textbf{Task coverage across the full generator spectrum.}
Figure~\ref{fig:s2f3} provides a task-coverage view across all 200 generators after ARC-TGI fine-tuning.
Although the frequency of successful tasks varies substantially across generators, solvable tasks appear throughout the full generator spectrum for all three models.
This suggests ARC-TGI captures diverse yet learnable transformation patterns without obvious idiosyncratic or model-specific anomalies.
Notably, Llama-3.1-8B solves tasks from more generators than the other two models, and the presence of overlapping solvable families across architectures supports the interpretation that generators reflect coherent rules that multiple model families can identify.

\section{Discussion}
\label{sec:discussion}

ARC-TGI reframes ARC(-AGI) evaluation from a static puzzle set into a controllable experimental platform: each original task becomes a resampleable \emph{task family} that yields fresh episodes with aligned reasoning traces and solver-facing code. This enables research directions that are difficult to study on fixed benchmarks.

\textbf{From leaderboard scores to diagnostic evaluation.}
Because ARC-TGI can sample matched task distributions, it supports analyses beyond aggregate accuracy.
One can (i) quantify robustness to nuisance variation by sweeping grid size, palette size, or distractor density; (ii) study within-family generalization by training on a subset of samples and evaluating on held-out samples from the same family; and (iii) probe cross-family transfer by training on ARC-AGI-1 families and evaluating on ARC-AGI-2 families.
A natural next step is to define standardized diagnostic slices (``stress tests'') over the generator suite.

\textbf{Induction vs.\ transduction under controlled variation.}
ARC-TGI supports both direct grid prediction and program synthesis and execution, enabling controlled comparisons of robustness under resampling.
Reasoning traces can serve as supervision and as a debugging signal when models fail, and the framework naturally supports hybrid systems that jointly produce a program and a trace or use traces for verification.

\textbf{Training data vs.\ evaluation data.}
ARC-TGI makes it straightforward to create large training corpora by sampling many tasks from training families.
At the same time, it highlights a conceptual distinction: sampling more tasks from the same family primarily increases \emph{within-family} diversity, whereas sampling across families increases \emph{rule} diversity.
Understanding how different learning systems benefit from these two sources of variation remains an open question.

\textbf{Human-in-the-loop generator authoring.}
Our development process combines LLM-assisted drafting with manual refinement and validation, reflecting a broader pattern: for benchmarks intended to remain human-solvable and cognitively grounded, fully automated generation is unlikely to suffice without strong constraint mechanisms and careful review.
Even with LLM acceleration, iteration and human checks remain essential, particularly for harder ARC-AGI-2 families.

\section{Conclusion}
\label{sec:conclusion}

ARC-TGI
reframes ARC(-AGI) evaluation from a fixed puzzle set into a controllable, resampleable platform by representing each task as a \emph{task family} with aligned reasoning traces and solver-facing code.
This enables diagnostic evaluation beyond leaderboard scores, including robustness sweeps over nuisance variation, within- vs.\ across-family generalization studies, and controlled comparisons between inductive program-based solvers and transductive direct-prediction approaches.
Across our empirical studies, ARC-TGI supports both in-distribution and out-of-distribution evaluation, revealing a persistent gap between learning and generalization in current LLMs while demonstrating consistent per-generator difficulty structure (e.g., Qwen3-30B at 21\% and Claude Sonnet 4.5 at 50\%).
ARC-TGI is paired with a human-in-the-loop authoring process that combines LLM-assisted drafting with manual refinement and validation to preserve human-solvable episodes.
We hope ARC-TGI serves as shared evaluation and data infrastructure for the community and supports standardized protocols, richer constraints, and improved validation methods for rigorous reasoning benchmarks.


\clearpage
\bibliographystyle{ACM-Reference-Format}
\bibliography{ref}

\appendix

\newpage
\section{Acknowledgments}
This work was supported by the German Federal Ministry of Education and Research (BMBF, SCADS22B) and the Saxon State Ministry for Science, Culture and Tourism (SMWK) by funding the competence center for Big Data and AI "ScaDS.AI Dresden/Leipzig" and TIB - Leibniz Information Centre for Science and Technology.

\section{ARC-TGI Example Tasks}
In order to generalise the original ARC-AGI tasks while also obtaining a wide distribution of samples, we introduce two different levels of variables: task variables and grid variables. Task variables are fixed within each individual sample and remain consistent across all training and test grids of that sample (in some specific tasks, the training grids share a fixed task variable while the test grid differs; these cases are handled within the task constraints). These variables operate at the level of the sample family and enable variation across different samples, for example in color, size, number of objects, position of objects, or orientation. In contrast, grid variables vary across the different examples within a single sample and are randomised in order to maintain high variation among the training and test grids. Together, these two types of variables support both generalisation and diversity.

\begin{figure*}[h!]
    \centering
    \includegraphics[width=0.6\textwidth]{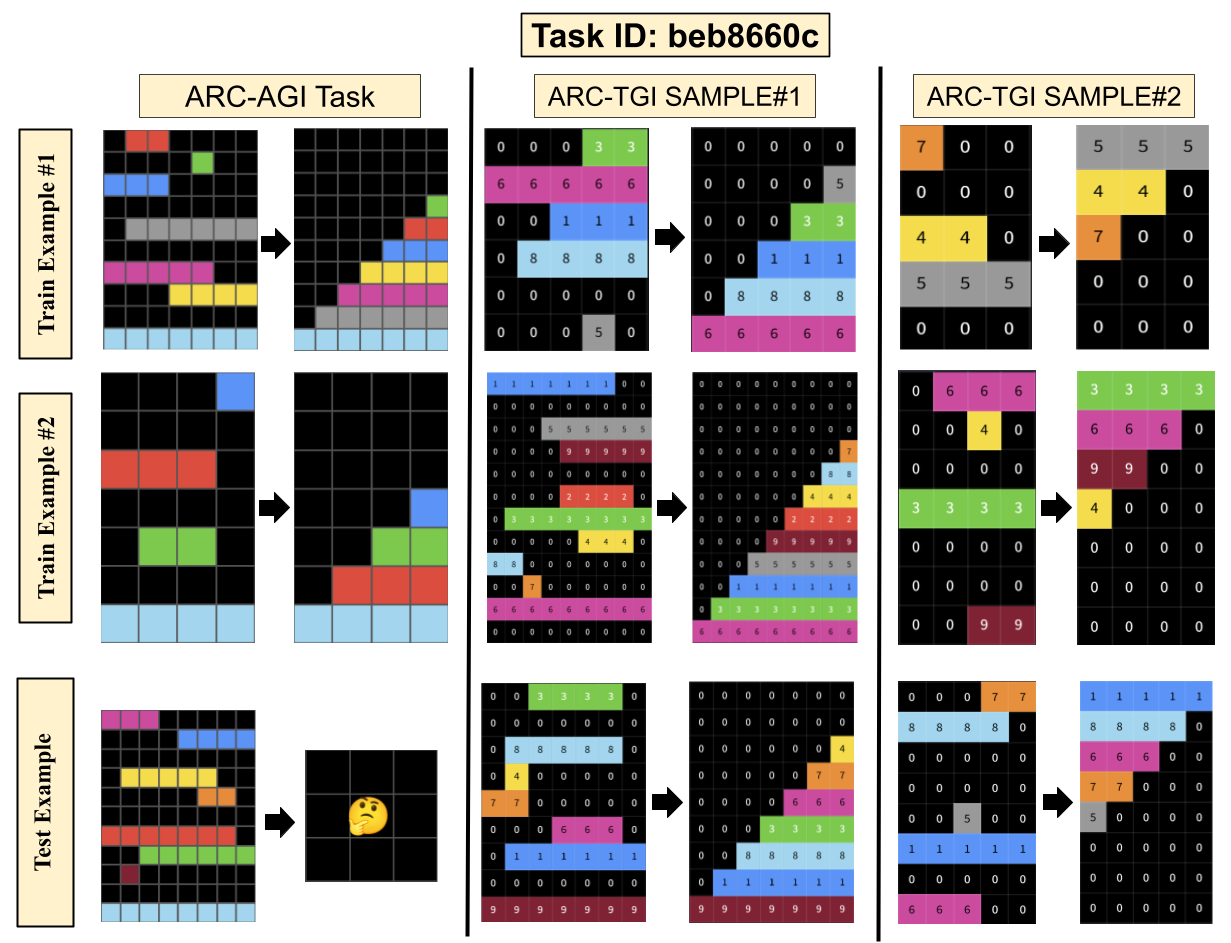}
    \caption{Within family task generalisation}
    \label{fig:task_generalisation}
\end{figure*}

In Figure~\ref{fig:task_generalisation}, the grid variables correspond to the number, length, and colours of the segments, which vary across the different training and test grids within a single sample. The task variables correspond to the vertical stacking direction of the segments (top or bottom) and their horizontal alignment (left or right), which are fixed across all grids within a sample. In the original ARC-AGI task, the coloured segments are stacked at the bottom and right-aligned. ARC-TGI, by means of the task variables, enables a generalisation of this task in which samples can be generated covering all combinations of vertical stacking and horizontal alignment, while preserving the underlying transformation logic.

\begin{figure*}[h!]
    \centering
    \includegraphics[width=0.6\textwidth]{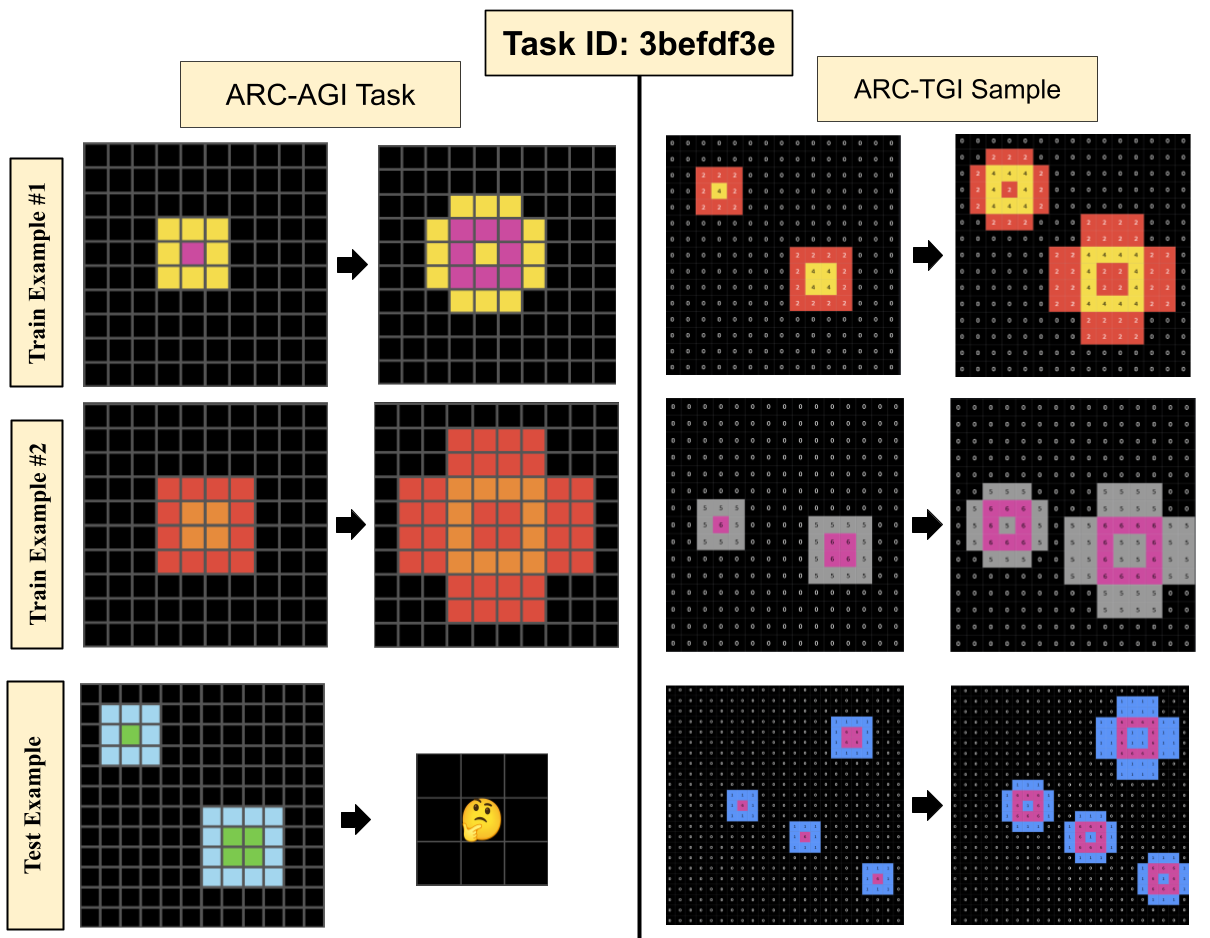}
    \caption{Train–test consistency}
    \label{fig:task-test-only}
\end{figure*}

Fig.\ref{fig:task-test-only} illustrates a case where care must be taken to avoid introducing test-only rules. To ensure this, each sample includes at least one training example containing a \(3 \times 3\) object and at least one training example containing a \(5 \times 5\) object. Since the transformation rule differs for each of them, omitting either size from the training examples would make the test example unsolvable. In addition, the number of objects in the training inputs always differs from that in the test input, reflecting the structure of the original ARC-AGI task.

\begin{figure*}[h!]
    \centering
    \includegraphics[width=0.6\textwidth]{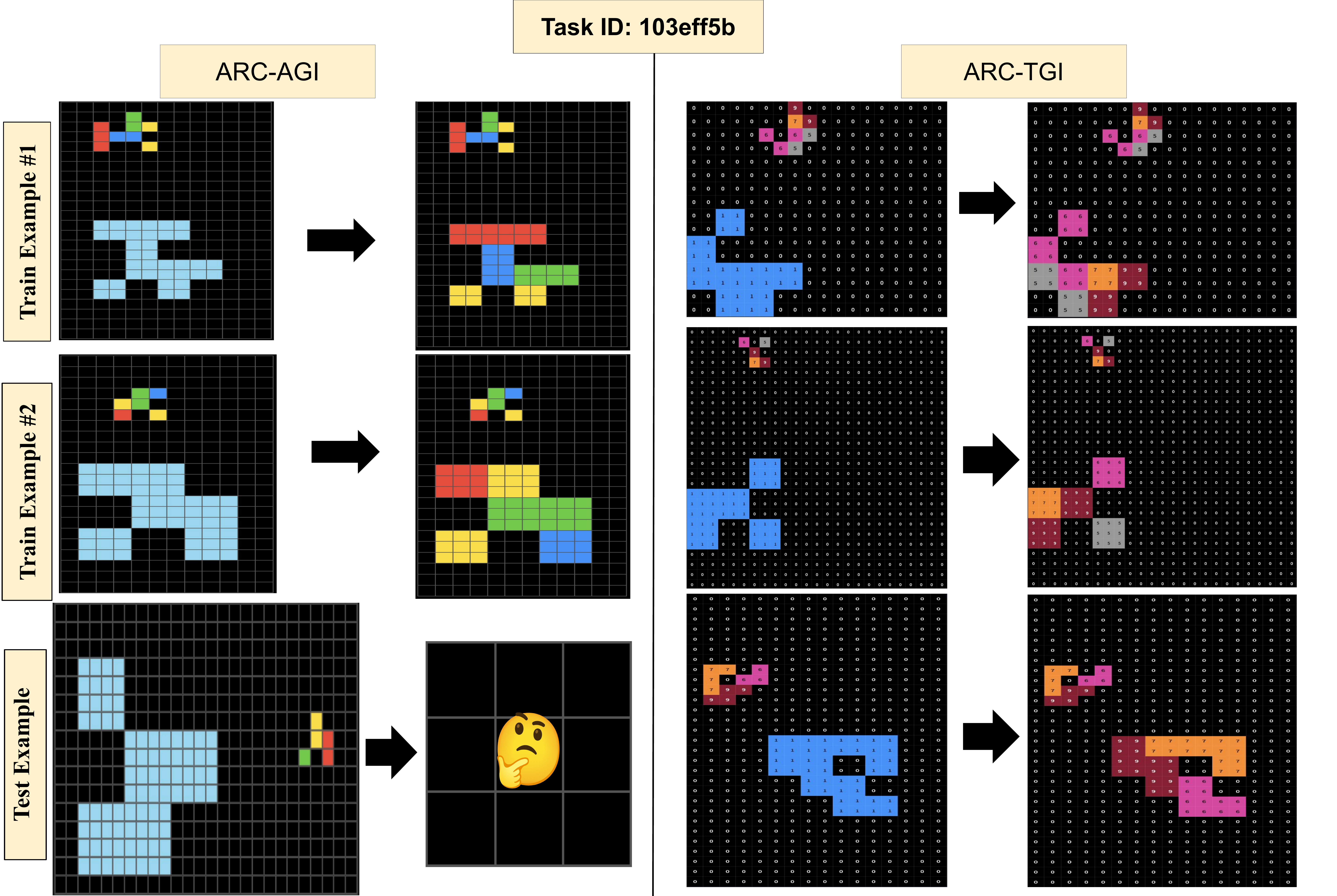}
    \caption{ARC-AGI task (Task ID: 103eff5b) and corresponding ARC-TGI sample.}
    \label{fig:task_103eff5b}
\end{figure*}

Fig.~\ref{fig:task_103eff5b} illustrates a case where ARC-TGI preserves task-level generalization constraints from the original ARC-AGI task (Task ID: 103eff5b). As in the original task, color semantics remain consistent between inputs and outputs across all examples, which in ARC-TGI is enforced by treating colors as task variables. Moreover, the train--test distinction is preserved: in the original ARC-AGI task, the number of colors used in the smaller object differs from that in the training examples, and the same controlled variation is maintained in the ARC-TGI sample.

\begin{figure*}[t]
  \centering
  \begin{subfigure}{0.8\textwidth}
    \includegraphics[width=\linewidth]{INPUT.pdf}
    \caption{Input grid-size heatmaps. Upper: original tasks. Lower: ARC-TGI samples (50 per generator).}
  \end{subfigure}

  \medskip

  \begin{subfigure}{0.8\textwidth}
    \includegraphics[width=\linewidth]{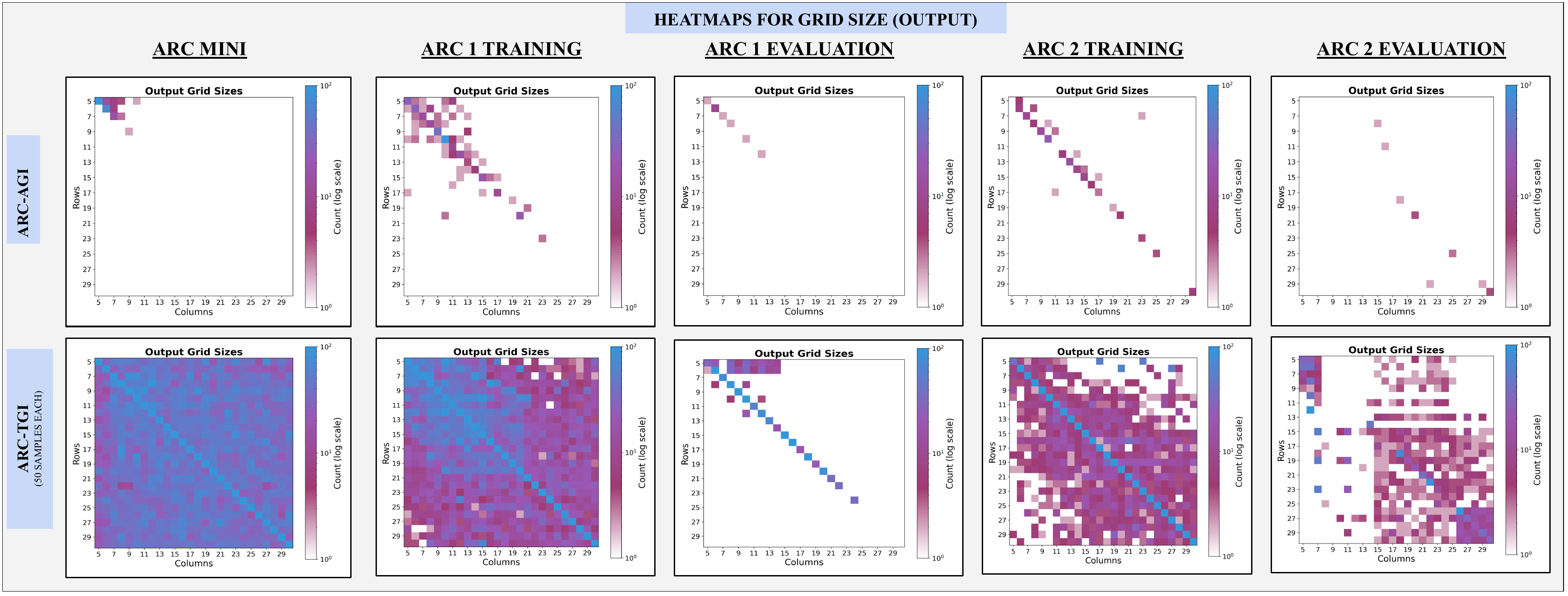}
    \caption{Output grid-size heatmaps. Upper: original tasks. Lower: ARC-TGI samples (50 per generator).}
  \end{subfigure}

  \label{fig:arcgen_heatmaps_combined}
\end{figure*}

\paragraph{Future work}
We see several high-leverage extensions.
First, ARC-TGI could support richer task-level constraint patterns through reusable templates for common disambiguation strategies (e.g., explicitly constructing counterexamples in the training set).
Second, suite-wide automatic validation and filtering could be strengthened with static checks for degenerate shortcuts and dynamic checks for ambiguity (e.g., detecting multiple consistent rules under simple hypothesis classes).
Third, broader adoption would benefit from standardized evaluation protocols, including community-agreed sampling budgets ($k$ per family), reporting conventions, and shared stress-test suites.
Fourth, lightweight human-solvability audits (time-to-solve and inter-annotator agreement on generated tasks) could provide an additional quality signal for generator maintenance.
Finally, concept tagging and search---via lightweight metadata or learned tags---could help users discover and evaluate targeted subsets of families (e.g., symmetry-heavy, object-centric, or multi-step compositional tasks).



\end{document}